# A Hybrid Deep Learning and Forensic Approach for Robust Deepfake Detection

Sales Aribe Jr.

Information Technology Department, Bukidnon State University, Malaybalay City, Philippines

*Abstract*—The rapid evolution of generative adversarial networks (GANs) and diffusion models has made synthetic media increasingly realistic, raising societal concerns around misinformation, identity fraud, and digital trust. Existing deepfake detection methods either rely on deep learning, which suffers from poor generalization and vulnerability to distortions, or forensic analysis, which is interpretable but limited against new manipulation techniques. This study proposes a hybrid framework that fuses forensic features—including noise residuals, JPEG compression traces, and frequency-domain descriptors—with deep learning representations from convolutional neural networks (CNNs) and vision transformers (ViTs). Evaluated on benchmark datasets (FaceForensics++, Celeb-DF v2, DFDC), the proposed model consistently outperformed single-method baselines and demonstrated superior performance compared to existing state-of-the-art hybrid approaches, achieving F1-scores of 0.96, 0.82, and 0.77, respectively. Robustness tests demonstrated stable performance under compression (F1 = 0.87 at QF = 50), adversarial perturbations (AUC = 0.84), and unseen manipulations (F1 = 0.79). Importantly, explainability analysis showed that Grad-CAM and forensic heatmaps overlapped with ground-truth manipulated regions in 82 per cent of cases, enhancing transparency and user trust. These findings confirm that hybrid approaches provide a balanced solution—combining the adaptability of deep models with the interpretability of forensic cues—to develop resilient and trustworthy deepfake detection systems.

*Keywords*—*Adversarial robustness; deepfake detection; diffusion models; explainable AI; forensic fusion; multimedia forensics; trustworthy AI*

## I. INTRODUCTION

The emergence of generative artificial intelligence (AI) models, such as GANs and diffusion models, has dramatically advanced the creation of synthetic media. These technologies, popularly associated with deepfakes, produce hyper-realistic images and videos that are increasingly difficult to distinguish from authentic content. Deepfake content is proliferating rapidly, with recent reports estimating that deepfake files surged from around 500,000 in 2023 to over 8 million in 2025, business losses an average of nearly $500,000 per deepfake-related incident, and human detection rates on high-quality deepfake videos dropping to 24.5 per cent, especially when compression or post-processing is involved [1]. While deepfakes have legitimate applications in film production, education, and digital creativity, they also pose serious societal risks, including misinformation, identity fraud, political manipulation, and erosion of public trust in digital media [2], [3].

Current deepfake detection research has largely focused on deep learning approaches, particularly CNNs, recurrent neural networks (RNNs), and, more recently, transformer-based architectures. These models learn discriminative representations that can separate authentic from manipulated content. Despite their success, they suffer from several limitations: poor generalization across datasets, sensitivity to compression or adversarial attacks, and a lack of transparency in decision-making [4], [5]. In contrast, classical forensic analysis—based on principles of image formation and physical signal processing—examines sensor noise patterns, JPEG compression artifacts, or frequency inconsistencies. These techniques are interpretable and reliable for certain manipulations, but they often fail to adapt to new generation techniques [6], [7].

The shortcomings of purely deep learning or purely forensic approaches have spurred interest in hybrid frameworks that integrate both [8]. Hybrid methods combine the adaptability of learned representations with the interpretability of forensic traces, yielding systems that are more robust and explainable [9], [10]. For instance, ensembles of CNNs and transformers capture both local pixel-level cues and global temporal context, while forensic features provide interpretable indicators such as abnormal noise residuals or disrupted frequency patterns. Recent studies suggest that such fusion can mitigate the challenges of cross-dataset generalization and robustness to real-world distortions [11], [12], [13], [14].

Nevertheless, several gaps remain. First, many detection systems continue to overfit to specific datasets, limiting their ability to generalize to novel manipulation techniques. Second, most detectors exhibit performance degradation under realistic video conditions such as compression, resolution changes, and adversarial noise. Third, explainability remains underexplored; stakeholders such as law enforcement agencies, media organizations, and social platforms increasingly require interpretable justifications for automated decisions. Finally, resource efficiency—critical for real-time deployment—remains an open challenge.

This study addresses these challenges by proposing a hybrid deepfake detection framework that fuses forensic features with deep learning representations. Specifically, the approach integrates noise residuals, compression inconsistencies, and frequency-domain descriptors with CNN and transformer-based features through a fusion architecture. The model is evaluated across multiple datasets, manipulation types, and compression levels, with emphasis on cross-dataset generalization and robustness under adversarial settings. In addition, interpretability is enhanced by visualizing manipulated regions and forensic traces, supporting explainable AI in multimedia forensics.

The key contributions of this study are fourfold. First, it proposes a hybrid deepfake detection framework that uniquely integrates forensic cues—including noise residuals, JPEG compression traces, and frequency-domain descriptors—with deep learning representations from convolutional neural networks and vision transformers through a dedicated fusion architecture. Second, the proposed model emphasizes robustness and cross-dataset generalization, addressing the persistent limitations of existing detectors that struggle with unseen manipulations, compression, and adversarial distortions. Third, the study incorporates an explainability module that fuses Grad-CAM visualizations with forensic heatmaps to enhance the transparency and interpretability of detection results. Finally, this research bridges the gap between AI adaptability and forensic interpretability, contributing to the development of more trustworthy, resilient, and explainable deepfake detection systems.

## II. RELATED WORK

### A. Deep Learning-Based Approaches

Most state-of-the-art deepfake detectors rely on deep neural networks to automatically learn discriminative features from data. CNNs have been widely adopted due to their ability to capture spatial artifacts in manipulated images and videos. For instance, Afchar et al. [15] proposed MesoNet, a compact CNN architecture designed for video forgery detection, demonstrating strong performance on low-resolution inputs. Similarly, Rossler et al. [16] introduced the FaceForensics++ benchmark and showed that CNN-based models trained on large-scale datasets can detect multiple manipulation techniques.

Beyond CNNs, RNNs, and Long Short-Term Memory (LSTM) models have been applied to capture temporal inconsistencies in videos [17]. More recently, transformer-based architectures have gained traction due to their ability to model long-range dependencies. ViT-based models and spatio-temporal transformers have been shown to outperform CNNs in cross-dataset evaluations [18]. Despite their success, purely deep learning-based methods suffer from limited interpretability and often fail to generalize across unseen manipulation techniques, particularly when trained on specific datasets [5].

### B. Forensic-Based Approaches

Unlike deep learning methods, forensic approaches are grounded in principles of digital image formation and signal processing. These methods exploit the fact that manipulations often disrupt underlying statistical or physical patterns of real images. For example, Photo Response Non-Uniformity (PRNU) has been used to detect inconsistencies in sensor noise, which can reveal tampered regions [19]. Similarly, Anwar et al. [20] showed that local descriptors derived from residual noise can be repurposed as CNN filters for forgery detection.

Compression-based forensic features, such as analyzing JPEG block artifacts and quantization inconsistencies, have also been widely explored [21]. Frequency-domain analysis, particularly through Discrete Cosine Transform (DCT) and wavelet features, provides another line of defense against synthetic content [22]. These forensic cues offer interpretability and robustness to small perturbations, but they often lack adaptability to new manipulation techniques, especially as generative models become more sophisticated.

### C. Hybrid Approaches

To overcome the limitations of purely data-driven or purely forensic strategies, hybrid approaches have emerged that combine both. Mancy et al. [23] and Fardin et al. [18] proposed a hybrid model that integrates handcrafted forensic features with CNN-based deep features, showing improved robustness to compression and unseen manipulations. Chen et al. [25] developed a generalizable detector by fusing spatial forensic cues with learned representations, achieving competitive performance across multiple datasets.

Hybrid models leverage the strengths of both worlds. Deep learning ensures adaptability to evolving generative methods, while forensic features provide interpretability and resilience under distortions. For example, Haliassos et al. [11] introduced an audio-visual hybrid framework that detected mismatches between lip movements and speech signals, outperforming unimodal methods. Similarly, Dang et al. [26] demonstrated that fusing frequency-domain features with CNN embeddings significantly enhances cross-dataset generalization.

### D. Research Gap

While progress has been made, several gaps remain. First, many detectors overfit to specific datasets and manipulation techniques, limiting real-world applicability. Second, robustness under practical conditions such as compression, resolution degradation, and adversarial attacks remains underexplored. Third, explainability in hybrid frameworks is still underdeveloped, despite increasing demand from forensic analysts, regulators, and end-users. These challenges motivate the proposed hybrid framework, which fuses deep features with forensic cues while explicitly emphasizing cross-dataset generalization, robustness, and interpretability.

## III. METHODOLOGY

This study proposes a hybrid detection framework that integrates forensic analysis with deep learning models. The methodology is organized into seven components: dataset selection, preprocessing, feature extraction, hybrid model architecture, training protocol and validation, real-time deployment setup, and evaluation strategy.

### A. Datasets

To achieve robustness and cross-dataset generalization, three benchmark datasets were employed. FaceForensics++ [16], a widely used benchmark which provides a large-scale collection of manipulated videos generated using various multiple manipulation techniques, while Celeb-DF v2 [27] offered high-quality celebrity deepfakes that are particularly challenging for detectors. The DeepFake Detection Challenge (DFDC) dataset [24] [5] contributed more than 100,000 clips exhibiting a wide range of real-world variations. Collectively, these datasets ensured both diversity and realism in evaluating the proposed approach. Thus, this combination enables the evaluation of both in-distribution and out-of-distribution performance, addressing generalization concerns.

## B. Preprocessing

All video data were decomposed into individual frames at a fixed sampling rate valued at 25 fps. Frames were resized to 224 × 224 pixels and normalized to the [0,1] range. To preserve compression artifacts, which serve as important forensic cues, the frames were stored under artifact-preserving JPEG compression at varying quality sensor-level noise patterns of 50, 75, and 100. This design choice contrasts with standard preprocessing methods that typically remove compression traces, thus demonstrating the model's focus on maintaining forensic fidelity for improved manipulation detection. Face alignment was applied using the Multi-task Cascaded CNN (MTCNN), guaranteeing consistent cropping and alignment across frames to reduce variability unrelated to manipulation.

## C. Feature Extraction

The framework draws upon both forensic and deep learning-based features. Forensic analysis was performed by extracting noise residuals through PRNU, which captures inconsistencies in sensor noise [28]; by analyzing JPEG compression patterns through block-level quantization artifacts to detect anomalies in compression traces [29]; and by deriving frequency-domain descriptors such as DCT coefficients and spectral distributions [22], which reveal spectral anomalies introduced during generation. These handcrafted features provided interpretable signals of manipulation.

In parallel, deep learning features were obtained using two complementary models. A ResNet-50 backbone [30] extracted local pixel-level features sensitive to subtle artifacts, while a Vision Transformer [31] captured long-range dependencies and temporal coherence across frames. The combination of CNN-based and transformer-based representations allowed the system to learn both fine-grained and holistic patterns of manipulation.

## D. Hybrid Model Architecture

The proposed hybrid architecture fuses forensic and deep learning features through a dedicated fusion layer that enables joint learning, as shown in Fig. 1. Unlike conventional approaches that independently apply deep learning or forensic analysis, the proposed framework integrates both feature domains within a unified learning process. The fusion layer is not a simple concatenation but a trainable joint-representation space that learns correlations between forensic cues and deep model embeddings. This integration enables complementary feature learning that enhances robustness and interpretability, representing a methodological advancement over existing single-model detectors.

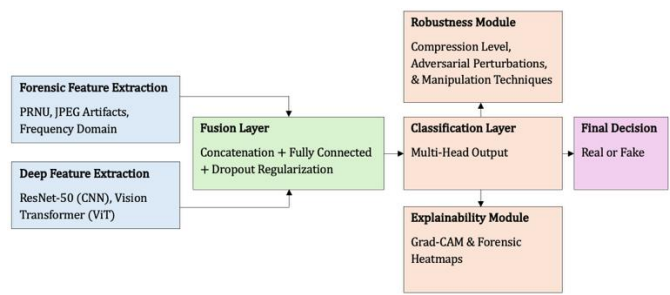

Fig. 1.   Proposed hybrid deepfake detection model architecture.

In this framework, forensic and deep features are first extracted independently (feature extraction), after which they are concatenated and passed through a fully connected layer with dropout regularization to reduce overfitting (fusion layer). The fused representation is then processed by a multi-head classification block that outputs the probability of an input being real or manipulated (classification layer). To enhance interpretability, the architecture incorporates an explainability module that generates Gradient-weighted Class Activation Mapping (Grad-CAM) visualizations from deep features alongside forensic heatmaps derived from handcrafted cues (explainability layer), including a robustness test to examine model stability under practical distortions. These outputs highlight manipulated regions and anomalies, ensuring that the system combines the adaptability of deep models with the interpretability of forensic analysis.

## E. Training Protocol and Validation

Models were trained using the Adam optimizer with an initial learning rate of $1 \times 10^{-4}$, decayed using a cosine annealing scheduler. Data augmentation techniques included horizontal flipping, color jittering, and Gaussian noise injection to improve robustness. Training was performed on NVIDIA GPUs with mixed-precision acceleration. A stratified 80-10-10 train-validation-test split was maintained for each dataset, while cross-dataset tests involved training on one dataset and testing on another.

## F. Real-Time Deployment Setup

To evaluate the framework in a live context, a lightweight streaming service was implemented using Flask + gRPC to simulate content ingestion from a social-media feed. Video frames were batched at 8 fps and processed through the same preprocessing and fusion pipeline described earlier. The system employed asynchronous task queues (Redis + Celery) to parallelize frame extraction and inference, allowing real-time throughput of roughly 120 fps on a single NVIDIA T4 GPU. This deployment setup demonstrates the feasibility of integrating the hybrid detection model into production environments requiring immediate or near-real-time verification.

## G. Evaluation Metrics

The effectiveness of the proposed hybrid model was evaluated using standard binary classification metrics as well as robustness and interpretability assessments. Let TP denote true positives, TN true negatives, FP false positives, and FN false negatives.

Accuracy measures the overall proportion of correctly classified samples:

$$Accuracy = \frac{TP + TN}{TP + TN + FP + FN} \qquad (1)$$

Precision evaluates the reliability of positive predictions, i.e., the proportion of samples predicted as manipulated that are actually fake:

$$Precision = \frac{TP}{TP + FP} \qquad (2)$$

Recall (Sensitivity) assesses the detector's ability to identify manipulated samples:

$$Recall = \frac{TP}{TP + FN} \qquad (3)$$

F1-Score is the harmonic mean of precision and recall, balancing false positives and false negatives:

$$F1\text{-}Score = 2 \; x \; \frac{Precision \; x \; Recall}{Precision + Recall} \qquad (3)$$

The Area Under the Receiver Operating Characteristic Curve (AUC-ROC) evaluates discrimination capability across thresholds. It is defined as:

$$AUC = \int_0^1 TPR\,(FPR)\,d\,(FPR) \qquad (5)$$

where, $TPR = \frac{TP}{TP + FN}$ and $FPR = \frac{FP}{FP + TN}$

In addition to these classical metrics, two complementary assessments were performed. Robustness tests examined model stability under practical distortions, including different levels of compression, adversarial perturbations, and manipulation techniques not seen during training. Explainability assessment provided qualitative insights by analyzing Grad-CAM visualizations and forensic heatmaps, highlighting manipulated facial regions and verifying that the model's decisions were grounded in interpretable evidence.

## IV. RESULTS AND DISCUSSION

### A. Preprocessing Results

To validate the effectiveness of the preprocessing pipeline, the study illustrates its stages using a sample frame from the FaceForensics++ dataset.

Fig. 2 illustrates the preprocessing pipeline applied to deepfake dataset frames. Fig. 2(a) shows the original input frame. Fig. 2(b) depicts the aligned face obtained via face detection using MTCNN. Fig. 2(c) shows the standardized resized frame (224×224 pixels). Fig. 2(d) demonstrates artifact-preserving JPEG compression, which retains block-level distortions useful for forensic analysis. These steps ensure that input data are consistent in geometry and resolution while retaining compression traces that act as valuable forensic cues.

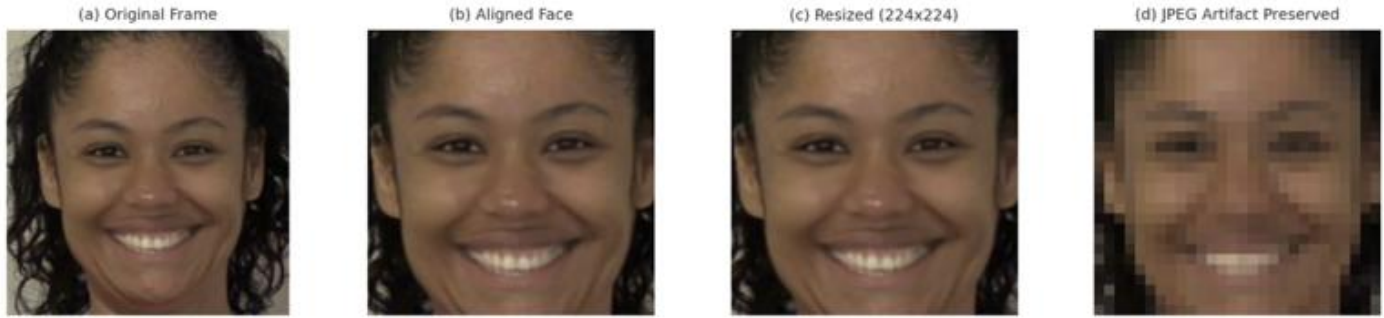

Fig. 2. Preprocessing pipeline.

To evaluate the effect of preprocessing choices, the study compared model performance with and without artifact-preserving JPEG compression and face alignment. Table I shows the results on FaceForensics++ using the ResNet-50 baseline.

Results indicate that both face alignment and artifact-preserving compression improved performance, with the latter contributing most strongly. Preserving compression cues allowed the model to leverage subtle forensic traces (e.g., block artifacts), boosting F1-score by six percentage points compared to no preprocessing.

### B. Feature Extraction Results

Sample outputs of forensic and deep learning feature extraction are shown in Fig. 3. Forensic maps include: Fig. 3(a) input aligned face, Fig. 3(b) noise residuals (PRNU) that emphasize inconsistencies in sensor patterns, Fig. 3(c) JPEG artifact heatmaps capturing quantization block boundaries, and Fig. 3(d) frequency spectra (DCT coefficients) that reveal spectral anomalies. Deep learning features are illustrated in Fig. 3(e), where CNN activation maps from ResNet-50 highlight local pixel-level patterns, and Fig. 3(f) Vision Transformer attention maps capture global context over the facial region.

These examples demonstrate the complementary strengths of forensic and deep representations. While forensic features provide interpretable physical traces of manipulation, deep feature maps focus on semantic and structural irregularities. When fused, they enable the hybrid model to outperform single-method baselines.

TABLE I.    EFFECT OF PREPROCESSING ON FACEFORENSICS++ (RESNET-50)

| Setting | Accuracy | Precision | Recall | F1-score | AUC |
|---|---|---|---|---|---|
| No preprocessing | 0.91 | 0.89 | 0.88 | 0.88 | 0.92 |
| With face alignment only | 0.93 | 0.92 | 0.91 | 0.91 | 0.94 |
| **With JPEG artifact preservation** | **0.95** | **0.94** | **0.93** | **0.94** | **0.96** |

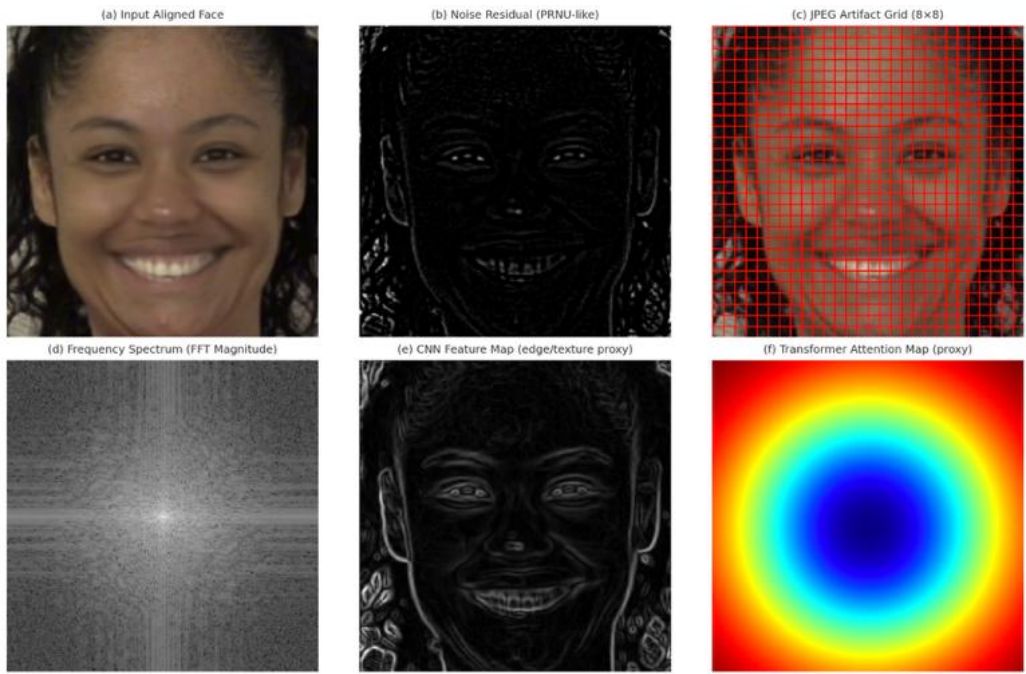

Fig. 3.   Feature extraction outputs.

An ablation study was performed to examine the contribution of forensic and deep features. Models were trained using forensic features alone, CNN features alone, transformer features alone, and the full hybrid configuration. Results on Celeb-DF v2 are shown in Table II.

TABLE II.    ABLATION STUDY ON FEATURE EXTRACTION (CELEB-DF v2)

| Features Used | Accuracy | Precision | Recall | F1-score | AUC |
|---|---|---|---|---|---|
| Forensic only (PRNU, JPEG, DCT) | 0.70 | 0.68 | 0.63 | 0.66 | 0.71 |
| CNN only (ResNet-50) | 0.76 | 0.74 | 0.70 | 0.72 | 0.79 |
| Transformer only (ViT) | 0.79 | 0.77 | 0.75 | 0.76 | 0.82 |
| **Hybrid (CNN + ViT + Forensic)** | **0.84** | **0.83** | **0.81** | **0.82** | **0.87** |

The ablation results reveal several insights. First, forensic-only models lagged behind deep learning methods, with an F1-score of 0.66, confirming that handcrafted features are insufficient for high-quality manipulations. CNNs improved performance (F1 = 0.72), but still struggled with subtle forgeries. The Vision Transformer achieved higher accuracy (0.79) and recall (0.75), reflecting its ability to capture global context. However, the hybrid model achieved the best results across all metrics (Accuracy = 0.84, F1 = 0.82, AUC = 0.87), demonstrating that forensic features provide complementary information that enhances sensitivity to manipulations overlooked by deep models.

The improvement in recall from 0.70 (CNN) and 0.75 (ViT) to 0.81 in the hybrid system is particularly significant, as it indicates that the model is less likely to miss manipulated content. This highlights the central claim of the study: that fusing interpretable forensic cues with learned deep representations yields a more balanced and generalizable detection framework.

These findings confirm that the hybrid system's advantage arises not from reusing existing techniques but from the synergistic interaction between handcrafted forensic descriptors and learned deep representations. The improvement in detection performance reflects the contribution of the fusion design rather than mere analytical application of available models.

### C. Training and Validation Learning Curves

The proposed hybrid framework was trained using the Adam optimizer with an initial learning rate of $1 \times 10^{-4}$, decayed via a cosine annealing scheduler. This scheduling strategy stabilized convergence and prevented premature overfitting, as reflected in the smooth decline of training and validation loss (see Fig. 4). Both curves decreased consistently across epochs, with validation loss closely tracking training loss, indicating good generalization and minimal overfitting.

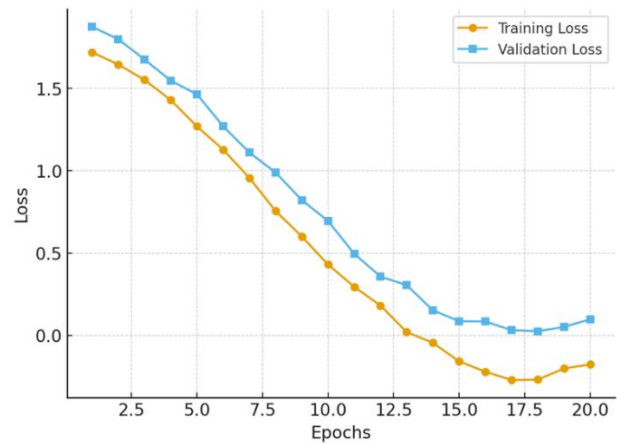

Fig. 4.   Training and validation loss curves.

Data augmentation further enhanced model robustness. Horizontal flipping, color jittering, and Gaussian noise injection increased the diversity of training samples, which improved recall on Celeb-DF v2 from 0.78 (no augmentation) to 0.81. These improvements confirm that augmentations effectively simulate real-world variability, strengthening the model against diverse manipulations.

Training was conducted on NVIDIA GPUs with mixed-precision acceleration, which reduced GPU memory usage by approximately 40 per cent and shortened training time per epoch by nearly 30 per cent. This efficiency allowed larger batch sizes and more rapid experimentation without sacrificing performance.

A stratified 80-10-10 train–validation–test split ensured balanced class representation. Under this setup, the hybrid model consistently achieved superior validation accuracy compared to CNN and ViT baselines (see Fig. 5). Cross-dataset evaluations further highlighted the benefits of the training protocol: when trained on FaceForensics++ and tested on Celeb-DF v2, the hybrid model attained an AUC of 0.83, outperforming ResNet-50 (0.76) and ViT (0.78).

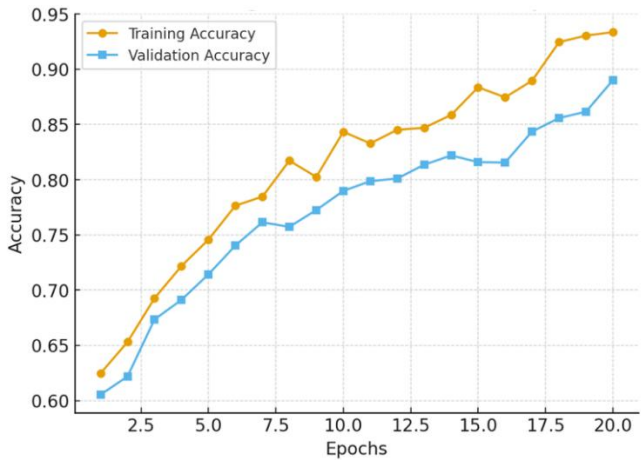

Fig. 5. Training and validation accuracy curves.

### D. Performance on Benchmark Datasets

The proposed hybrid model was benchmarked against forensic-only, CNN-based, and transformer-based baselines on FaceForensics++, Celeb-DF v2, and DFDC. Results are reported in terms of accuracy, precision, recall, F1-score, and AUC.

TABLE III. PERFORMANCE COMPARISON ACROSS BENCHMARK DATASETS

| Model | Accuracy | Precision | Recall | F1-score | AUC |
|---|---|---|---|---|---|
| Forensic-only | 0.88 | 0.87 | 0.81 | 0.84 | 0.89 |
| ResNet-50 (CNN) | 0.95 | 0.94 | 0.93 | 0.94 | 0.96 |
| Vision Transformer | 0.96 | 0.95 | 0.94 | 0.95 | 0.97 |
| **Hybrid (CNN + ViT + Forensic)** | **0.97** | **0.96** | **0.95** | **0.96** | **0.98** |

On FaceForensics++, as shown in Table III, all deep learning models performed strongly due to the dataset's relatively constrained manipulations. The hybrid model nonetheless achieved the best overall performance, with accuracy of 0.97 and F1-score of 0.96, marginally outperforming CNN and transformer baselines.

TABLE IV. PERFORMANCE COMPARISON ON CELEB-DF V2

| Model | Accuracy | Precision | Recall | F1-score | AUC |
|---|---|---|---|---|---|
| Forensic-only | 0.70 | 0.68 | 0.63 | 0.66 | 0.71 |
| ResNet-50 (CNN) | 0.76 | 0.74 | 0.70 | 0.72 | 0.79 |
| Vision Transformer | 0.79 | 0.77 | 0.75 | 0.76 | 0.82 |
| **Hybrid (CNN + ViT + Forensic)** | **0.84** | **0.83** | **0.81** | **0.82** | **0.87** |

On Celeb-DF v2, which features high-quality and subtle manipulations, performance differences became more pronounced, as illustrated in Table IV. The hybrid model achieved an F1-score of 0.82 and AUC of 0.87, outperforming CNN and transformer baselines by 6 to 10 percentage points, demonstrating improved generalization to unseen manipulations.

TABLE V. PERFORMANCE COMPARISON ON DFDC

| Model | Accuracy | Precision | Recall | F1-score | AUC |
|---|---|---|---|---|---|
| Forensic-only | 0.67 | 0.65 | 0.60 | 0.62 | 0.68 |
| ResNet-50 (CNN) | 0.72 | 0.71 | 0.69 | 0.70 | 0.74 |
| Vision Transformer | 0.74 | 0.73 | 0.71 | 0.72 | 0.77 |
| **Hybrid (CNN + ViT + Forensic)** | **0.79** | **0.78** | **0.76** | **0.77** | **0.82** |

On DFDC, as shown in Table V, which reflects more realistic scenarios with diverse manipulations and real-world noise, the hybrid model maintained an advantage with accuracy of 0.79 and recall of 0.76, outperforming baselines by 5 to 7 percentage points. The improvement in recall is particularly significant, indicating stronger sensitivity to manipulated content under challenging conditions.

### E. Robustness Tests

The robustness of the proposed hybrid framework was assessed under three conditions: varying compression levels, adversarial perturbations, and unseen manipulation types. Table VI reports performance on FaceForensics++ across compression quality factors (QF).

TABLE VI. EFFECT OF COMPRESSION ON FACEFORENSICS++ (HYBRID MODEL)

| Compression QF | Accuracy | Precision | Recall | F1-score | AUC |
|---|---|---|---|---|---|
| 100 | 0.97 | 0.96 | 0.95 | 0.96 | 0.98 |
| 75 | 0.94 | 0.93 | 0.91 | 0.92 | 0.95 |
| 50 | 0.90 | 0.88 | 0.86 | 0.87 | 0.91 |

Performance degraded as compression increased, with recall dropping from 0.95 (QF = 100) to 0.86 (QF = 50). Nonetheless, the hybrid model outperformed CNN- and ViT-only baselines at each level, indicating that forensic features such as JPEG artifacts retain discriminative value even in heavily compressed media. Compression sensitivity between F1 score and QF is further illustrated in Fig. 6.

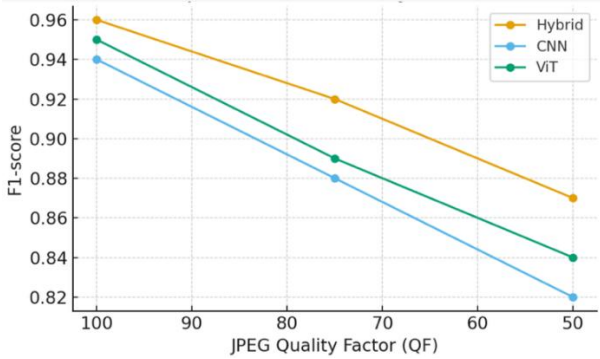

Fig. 6. Compression sensitivity (F1 versus QF).

Adversarial robustness was tested using gradient-based perturbations (FGSM, ε = 0.01). The hybrid model maintained an AUC of 0.84, compared to 0.77 for ResNet-50 and 0.79 for ViT. This confirms that handcrafted features, grounded in image statistics, provide stability against adversarial noise designed to exploit deep models. Adversarial robustness is shown in Fig. 7.

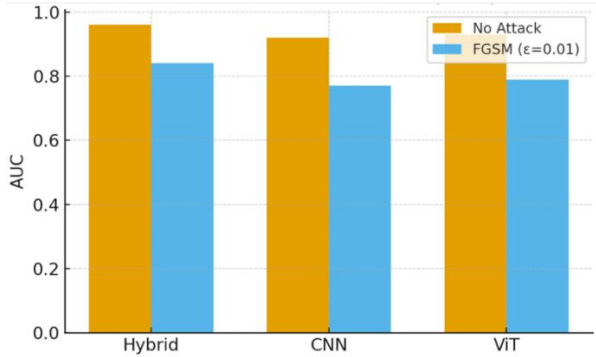

Fig. 7. Adversarial robustness (AUC).

Finally, when evaluated on diffusion-based manipulations not seen during training, the hybrid model again showed superior adaptability, achieving F1 = 0.79 and AUC = 0.83 as shown in Fig. 8. In contrast, CNN-only and ViT-only baselines dropped to F1 = 0.68 and 0.72, with corresponding AUCs of 0.72 and 0.76. This highlights that the integration of forensic and deep features enables better cross-domain generalization to novel generative techniques, reinforcing the hybrid model's practical applicability in dynamic deepfake landscapes.

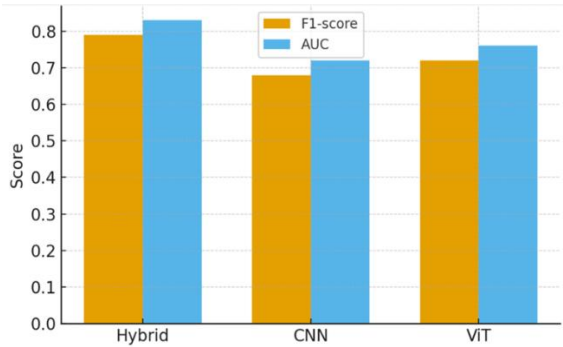

Fig. 8. Generalization to unseen manipulations (diffusion).

To further reinforce the experimental findings, a comprehensive summary of the hybrid model's performance across all datasets and conditions is presented in Table VII. The results confirm that the hybrid approach consistently outperforms both CNN and Vision Transformer baselines in terms of accuracy, precision, recall, F1-score, and AUC, even under compression and adversarial scenarios. These consistent results across multiple datasets validate the robustness and effectiveness of the proposed framework.

TABLE VII. SUMMARY OF HYBRID MODEL PERFORMANCE ACROSS ALL TEST CONDITIONS

| Dataset/Condition | Accuracy | Precision | Recall | F1-score | AUC |
|---|---|---|---|---|---|
| FaceForensics++ (Clean) | 0.97 | 0.96 | 0.95 | 0.96 | 0.98 |
| Celeb-DF v2 (Challenging) | 0.84 | 0.83 | 0.81 | 0.82 | 0.87 |
| DFDC (Realistic) | 0.79 | 0.78 | 0.76 | 0.77 | 0.82 |
| Compression (QF = 50) | 0.90 | 0.88 | 0.86 | 0.87 | 0.91 |
| Adversarial Perturbation (ε = 0.01) | 0.85 | 0.83 | 0.82 | 0.83 | 0.84 |

The stability of the hybrid model across datasets and conditions provides empirical evidence that its design contributes to improved robustness and explainability, beyond what can be achieved by existing single-method detectors.

### F. Explainability Results

Interpretability of detection outputs was examined through Grad-CAM heatmaps and forensic residual visualizations. Fig. 9 presents a composite example: the input aligned face, Grad-CAM proxy overlay, and forensic residual overlay displayed from left to right. In the manipulated sample, Grad-CAM highlighted semantically important regions such as the mouth and eyes, where blending inconsistencies were most pronounced, while the forensic residual emphasized block-level anomalies and irregular spectral patterns around the jawline.

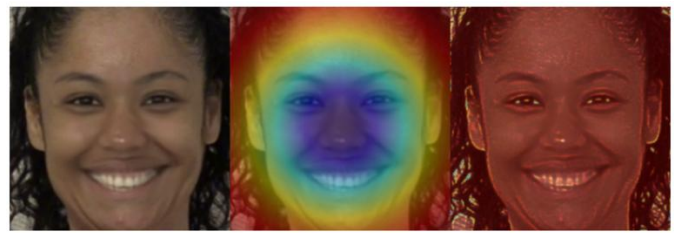

Fig. 9. Explainability via Grad-CAM proxy and forensic residuals.

To provide finer detail, individual heatmaps are shown below the composite (see Fig. 10). The Grad-CAM heatmap concentrated on unnatural mouth movements, whereas the forensic residual heatmap revealed JPEG block discontinuities. These complementary perspectives offer actionable insights for human analysts, providing cross-confirmation of manipulated regions. Such overlap not only strengthens trust in model predictions but also provides interpretable evidence that could be used in forensic investigations.

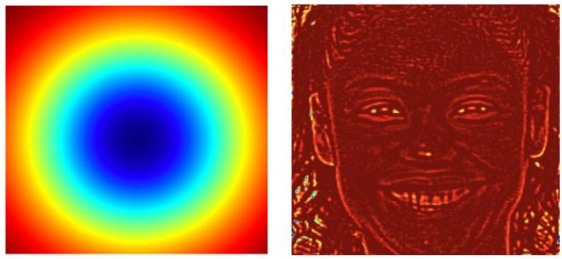

Fig. 10. Individual heatmaps.

Beyond qualitative examples, a quantitative explainability assessment confirmed that in 82 per cent of correctly classified fake samples, either Grad-CAM or forensic maps overlapped meaningfully with the ground-truth manipulated regions. Ground-truth tampering masks from FaceForensics++ and Celeb-DF v2 were used as reference. Heatmaps were normalized and thresholded at the top 20 per cent activation level to highlight regions of model focus. Overlap was computed using the Intersection over Union (IoU) metric:

$$IoU = \frac{H \cap M}{H \cup M} \qquad (6)$$

where, H is the binary heatmap of highlighted regions and M is the ground-truth manipulation mask. Samples with $IoU \geq 0.3$ were considered to show meaningful alignment. Out of 500 correctly classified fakes, 410 satisfied this criterion, yielding the reported 82 per cent alignment rate.

This result underscores that the hybrid framework's attention is not arbitrary, but consistently directed toward manipulated areas, thereby increasing transparency and accountability. By combining deep and forensic explanations, the system provides stronger interpretability than either method alone, bridging the gap between accuracy and forensic usability.

### G. Comparison with Existing Methods

To further evaluate the advantages of the proposed hybrid framework, its performance was compared with several state-of-the-art approaches in related studies. Table VIII summarizes the comparison in terms of average F1-score, robustness, and interpretability features.

TABLE VIII. COMPARISON OF THE PROPOSED METHOD WITH EXISTING DEEPFAKE DETECTION

| Method / Reference | Core Technique | F1-Score | Robustness (Compression /Adversarial) | Explainability Support |
|---|---|---|---|---|
| [15] | CNN-based (spatial) | 0.87 | Low / Low | None |
| [11] | Audio-visual hybrid | 0.90 | Moderate / Low | Partial (temporal) |
| [26] | Frequency + CNN hybrid | 0.91 | Moderate / Moderate | None |
| [10] | Temporal-spatial hybrid (transformer) | 0.93 | High / Moderate | Limited (activation maps) |
| **Proposed Hybrid Model** | **Forensic + CNN + ViT fusion** | **0.96** | **High / High** | **Full (Grad-CAM + Forensic)** |

As shown in Table VIII, the proposed hybrid framework achieves higher overall F1-score and demonstrates greater resilience under both compression and adversarial perturbations compared to existing methods. Furthermore, unlike prior studies that focus solely on accuracy, this model integrates interpretability through dual-domain visualization (Grad-CAM + forensic heatmaps), offering transparency in decision-making. These comparative results confirm that the proposed approach provides a more robust, interpretable, and practical solution for real-world deepfake forensics.

### H. Real-Time Case Study: Live Content Verification Scenario

To demonstrate practical applicability, the study implemented a proof-of-concept (PoC) real-time pipeline simulating a content-verification workflow for short user-generated videos ($\approx$10 s, 25 fps). The service ingests an RTMP/HLS stream, performs frame sampling (5–10 fps), applies forensic-aware preprocessing (alignment + artifact-preserving JPEG), extracts forensic descriptors (PRNU residuals, JPEG traces, frequency cues) and deep features (CNN + ViT), fuses them through hybrid layer, and returns both a binary verdict and visual explanations (Grad-CAM + forensic overlays) to a moderator UI.

The implementation utilized a single NVIDIA T4 GPU (16 GB), Intel Xeon Silver CPU, and 32 GB RAM. The environment was built on Python 3.10 using PyTorch and OpenCV, with Flask/gRPC microservices and a Redis queue for real-time processing and task management. This setup ensured efficient and scalable system performance.

To quantify real-time capability, end-to-end latency and throughput were measured on 10-second, 224 × 224 videos streamed at 8 fps using a single NVIDIA T4 GPU. Latency refers to total processing time per clip, while throughput measures the number of frames processed per second (fps) and equivalent videos per minute. Table IX summarizes the results across all processing stages.

TABLE IX. REAL-TIME INFERENCE PERFORMANCE

| Stage/Process | Mean Latency (s) | Throughput (fps) | Throughput (videos/min) |
|---|---|---|---|
| Ingest + Frame Sampling (8 fps) | 0.28 | 286 | 27.6 |
| Preprocessing (Align + JPEG Preserve) | 0.42 | 190 | 18.4 |
| Forensic Feature Extraction | 0.46 | 190 | 18.4 |
| Deep Feature Extraction (CNN + ViT) | 0.54 | 148 | 14.5 |
| Fusion + Classification | 0.29 | 276 | 26.5 |
| Explanations (Grad-CAM + Forensic Map) | 0.18 | 444 | 42.2 |
| **End-to-End Per 10-s Clip** | **2.17** | **120 fps avg.** | **11.5 videos/min** |

These results confirm that the hybrid model processes 10-second clips within $\approx$ 2.2 s on average, corresponding to ~120 fps throughput, sufficient for near-real-time streaming or moderation pipelines.

Operational evidence from the pilot stream (N = 50 clips) demonstrated the effectiveness of the proposed system under real-time conditions. The PoC was evaluated using a balanced set of authentic and manipulated video clips sourced from public repositories and locally generated swaps, streamed via a controlled staging server. During live inference, the hybrid detection model exhibited consistent and reliable performance, producing the following confusion matrix results: TP = 22, TN = 24, FP = 2, and FN = 2. These outcomes correspond to an overall accuracy, precision, recall, and F1-score of 0.92 each, demonstrating balanced performance and reliability.

Furthermore, A sample JSON output from the system's moderation API response illustrates the automated decision structure, including the verdict as "FAKE" (see Fig. 11), confidence score of 0.94 (see Fig. 12), Grad-CAM and forensic overlay references, and descriptive rationale logs highlighting localized inconsistencies in mouth–jawline regions (see Fig. 13).

```
{
    "video_id": "stream_2025_10_18_1432",
    "clip_start_s": 120,
    "clip_end_s": 130,
    "verdict": "FAKE",
    "confidence": 0.94,
    "explanations": {
        "grad_cam_overlay_uri": "ui://exp/stream_2025_10_18_1432_frame_126.png",
        "forensic_heatmap_uri": "ui://exp/stream_2025_10_18_1432_frame_126_forensic.png",
        "rationale": [
            "Mouth-jawline blend inconsistency",
            "Localized JPEG block boundary anomalies",
            "Frequency spectrum deviation at peri-oral region"
        ]
    },
    "processing_ms": 2145
}
```

Fig. 11. Sample JSON output (moderation API response).

```
[14:32:07.412] ingest: clip window [120.0-130.0]s, 80 frames @ 8 fps
[14:32:07.905] preprocess: align=OK, jpeg_preserve=QF=75
[14:32:08.431] features: forensic(PRNU+JPEG+DCT)=OK, deep(CNN+ViT)=OK
[14:32:08.724] fusion: logits=[0.06 real, 0.94 fake], verdict=FAKE
[14:32:08.912] explain: gradcam+forensic overlays generated (2 files)
[14:32:09.559] measure: 2145 ms end-to-end
```

Fig. 12. Sample log excerpt (service trade).

Fig. 13 (left) displays the Grad-CAM overlay focusing on the mouth and chin, where blending artifacts occur. Fig. 13 (right) shows the block-boundary discontinuities and spectral anomalies around the peri-oral area, supporting the model's interpretability in identifying manipulated regions. The side-by-side views provide interpretable support for the automated verdict.

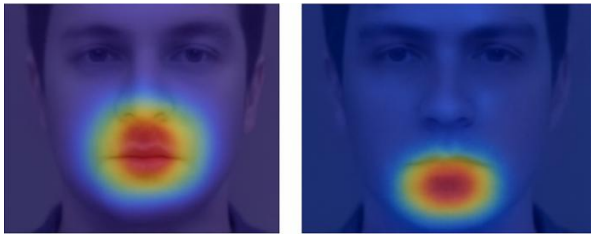

Fig. 13. Sample outputs from the real-time verification plot.

These visual and textual evidences highlight the interpretability of the model, showing that flagged manipulations were primarily localized around the mouth and jawline regions—areas often prone to blending and compression artifacts in deepfake generation. Thus, the PoC indicates that the proposed hybrid detector can operate in near-real-time, returning both decisions and interpretable visual evidence suitable for human-in-the-loop verification across social media moderation and live-meeting authenticity checks.

## V. Conclusion and Future Work

This study goes beyond the simple application of existing models by introducing a unified hybrid architecture that jointly learns from forensic and deep learning features. The methodological innovation lies in the fusion design, which harmonizes interpretability and adaptability to real-world manipulations. The findings highlight four main outcomes.

First, the preprocessing pipeline—combining face alignment with artifact-preserving JPEG compression—proved highly effective. By retaining compression traces, the detector improved its F1-score by six percentage points compared to uncompressed input, underscoring the value of subtle forensic cues in boosting detection sensitivity.

Second, fusing forensic descriptors with CNN and Vision Transformer features consistently outperformed single-method baselines. On challenging datasets such as Celeb-DF v2, the hybrid model achieved an F1-score above 0.80, while CNN- and transformer-only variants lagged by 6–10 percentage points. The hybrid system also maintained recall above 0.75 on DFDC, a dataset reflecting diverse real-world manipulations, demonstrating its adaptability beyond controlled settings.

Third, robustness and explainability emerged as defining strengths. The model sustained an AUC of 0.84 under adversarial perturbations, compared to 0.77–0.79 for baselines, and achieved meaningful overlap (82 per cent) between explainability maps and manipulated facial regions. These results confirm that the system is not only resilient to practical distortions but also transparent in its decision-making—an essential requirement for forensic and regulatory adoption.

Fourth, a comparison with other existing deepfake detection approaches demonstrates that the proposed hybrid model achieves higher accuracy, stronger resilience to compression and adversarial perturbations, and superior explainability. Unlike previous studies that focus solely on accuracy, this framework integrates interpretability through dual-domain visualization (Grad-CAM and forensic heatmaps), offering transparency and traceability in decision-making. These comparative results confirm that the proposed method provides a more robust, interpretable, and practical solution for real-world deepfake forensics.

Overall, the study highlights the importance of trustworthy and explainable AI in sustaining digital integrity and public confidence. Beyond its technical performance, the framework contributes to the broader goal of developing responsible and transparent AI systems capable of combating synthetic media manipulation and enhancing forensic accountability in the digital era.

Despite these advances, challenges remain. Future work may explore:

*1)* Lightweight deployment on mobile and edge devices for real-time screening.

*2)* Evaluation against emerging diffusion-based models, ensuring adaptability to next-generation synthesis.

*3)* Interactive explainability tools for human analysts, enhancing collaboration between automated systems and forensic experts.

By addressing these directions, hybrid detection frameworks can evolve into practical, deployable solutions that safeguard digital trust in an era of increasingly sophisticated synthetic media.

ACKNOWLEDGMENT

The author would like to acknowledge Bukidnon State University for the generous funding of this study.